\documentclass{article}

\usepackage{spconf,amsmath,epsfig}
\usepackage{hyperref}
\usepackage{bbm}
\usepackage{multirow} 
\usepackage{graphicx}
\usepackage[position=top]{subfig}
\usepackage{color}
\usepackage[dvipsnames]{xcolor}
\usepackage{colortbl}
\def\x{{\mathbf x}}
\def\X{{\mathbf X}}
\def\z{{\mathbf z}}
\def\Z{{\mathbf Z}}

\def\y{{\mathbf y}}
\def\Y{{\mathbf Y}}
\def\W{{\mathbf W}}
\def\Xt{{\tilde{{\mathbf X}}}}
\def\K{{\mathbf K}}
\def\k{{\mathbf k}}
\def\I{{\mathbf I}}

\def\H{\mathbf{H}}
\def\Real{{\mathbbm{R}}}
\def\Comp{{\mathbbm{C}}}
\newcommand{\imag}{\mathbbm{i}}
\newcommand{\red}[1]{\textcolor[rgb]{0.8,0,0}{#1}}       

\hypersetup{
    bookmarks=true,         
    unicode=false,          
    pdftoolbar=true,        
    pdfmenubar=true,        
    pdffitwindow=false,     
    pdfstartview={FitH},    
    pdftitle={},    
    pdfauthor={Camps-Valls},     
    pdfsubject={},   
    pdfcreator={Camps-Valls},   
    pdfproducer={Camps-Valls}, 
    pdfkeywords={keyword1} {key2} {key3}, 
    pdfnewwindow=true,      
    colorlinks=true,       
    linkcolor=blue,          
    citecolor=blue,        
    filecolor=blue,      
    urlcolor=blue           
}

\title{Nonlinear Cook distance for Anomalous Change Detection}

\name{Jos\'e A. Padr\'on Hidalgo$^{1}$, Adri\'an P\'erez-Suay$^{1}$, Fatih Nar$^{2}$, Gustau Camps-Valls$^{1}$
	\thanks{The research was funded by the European Research Council (ERC) under the ERC-CoG-2014 SEDAL project (grant agreement 647423), and by the Spanish Ministry of Economy and Competitiveness (MINECO) and European Regional Development Fund (ERDF) through the project TIN2015-64210-R and by the Scientific and Technical Research Council of Turkey (TUBITAK) with grant number TUBITAK-BIDEB-2219.}
}
\address{$^{1}$Image Processing Laboratory (IPL), Universitat de Val\`encia, Val\`encia, Spain\\
$^{2}$Konya Food and Agriculture University, Konya, Turkey}

\begin{document}

\maketitle
\begin{abstract}
In this work we propose a method to find anomalous changes in remote sensing images based on the chronochrome approach. 
A regressor between images is used to discover the most {\em influential points} in the observed data. Typically, the pixels with largest residuals are decided to be anomalous changes. In order to find the anomalous pixels we consider the Cook distance and propose its nonlinear extension using random Fourier features as an efficient nonlinear measure of impact. Good empirical performance is shown over different multispectral images both visually and quantitatively evaluated with ROC curves. 
\end{abstract}


\section{Introduction}
\label{sec:introduction}

Change detection~\cite{Lu04,Kwon03} is one of the most active fields in Remote Sensing and Earth Observation. Essentially, change detection boils down to identification of the abnormality buried in the background. An interesting broader problem is that of anomalous change detection (ACD): such setting differs from standard change detection in that the goal is to find anomalous or rare changes that occurred between two images, not simply pervasive changes.  

The interest to find anomalous changes in scenes is very broad, and many methods have been proposed in the literature, ranging from equalization-based approaches that rely on whitening principles~\cite{Mayer03}, to multivariate methods that extract distinct features out of the change (difference) image~\cite{Arenas13} and that reinforce directions in feature spaces associated with noisy or rare events~\cite{Green88,Nielsen98}, as well as regression-based approaches like in the chronochrome~\cite{Schaum97}, where a regressor tries to approximate the next incoming image and the big residuals are associated with anomalies. 

In this paper we elaborate on the chronochrome approach~\cite{Gustau},~\cite{Theiler10}. Two important choices need to be made: the regression algorithm and the criterion to evaluate residuals. Very often a simple linear model is used in the literature, but nonlinear algorithms like neural networks or kernel machines may offer better fitted models and thus less ambiguous residuals. On the other hand, there are many ways of looking at the residuals~\cite{REGRESSION1}, 
but more important than residuals inspection is the problem of translating residuals into anomalies. One could trivially associate big errors to anomalies, but a robust/hypothesis statistic test has to be defined. This is a field widely studied in descriptive statistics. 

A well-established method in statistics that summarizes the impact or influence of individual points in a regression model is the Cook's distance~\cite{Cook77}. Essentially the Cook's distance is defined as the sum of all changes in the regression model when a particular observation is removed. To our knowledge, the distance has not been used yet in image change detection in a chronochrome setting. Application of the Cook's distance in this scenario has to face, however, two important problems: 1) the distance is typically computed after application of a linear regression model, hence non-linear impact/changes can not be identified; and 2) the Cook's distance statistic for out-of-sample points requires the evaluation of as many models as test samples, which makes the technique very costly. In this paper, we provide a remedy to both problems. First, we extend the linear regression model to work with kernel functions hence accounting for non-linearities. However, noting the large computational cost involved, we propose the approximation of the kernels with explicit projections  on random Fourier features~\cite{Rahimi07nips}. The method is simple, computationally very efficient in both memory and processing costs, and achieves improved detection compared to standard approaches. We show results in a set of four change detection problems with pairs of multispectral images acquired by different sensors (Quickbird, Sentinel-2) and involving different changes of interest (floods, wildfires, urbanization). 


The remainder of the paper is organized as follows.
First, section \ref{sec:cook} fixes notation, introduces the Cook's distance, and briefly reviews the concept of influential points and leveraging in standard statistics. Section \ref{sec:experiments} presents the performance of the proposed randomized Cook's chronocrome method for anomaly change detection. 
Finally, we conclude in section \ref{sec:conclusion} with some remarks and prospective future work.

\section{Kernelized Cook's distance}
\label{sec:cook}

\subsection{Notation and the chronochrome approach}

Let us define two consecutive $d$-bands multispectral images in matrix form $\X,\Y\in\Real^{n\times d}$ composed of $n$ pixels $\x_i, \y_i\in\Real^d$, $i=1,\ldots,n$. Let us assume that a set of (anomalous) changes have occurred in between, and that such changes do not alter the image distribution significantly. 
The `chronochrome' approach~\cite{Chronochrome} departs from this idea to fit a function able to predict the second image $\Y$ from the first one $\X$, and decide that a point is anomalous if, for instance, the corresponding residual is significantly large. 
The prediction function $f: \x\to \y$ is learned (fitted) from the observations. 
The task is now to assess the significance of the obtained residuals, $\mathbf{e} = \y-\hat{\y}$. 

\subsection{Cook's distance}
\label{sec:Cook_distance}

Estimating {\em influential points} from models is a challenging problem in statistics. The aim is to find which elements from the sample set are more relevant to the statistical model. 
The Cook's distance~\cite{Cook77} is a standard tool from descriptive statistics to assess the influence of individual observations on the fitted model. 


The standard Cook's distance assumes a linear model for prediction of the second image from the first one, i.e. $\hat{\Y}=\Xt\W$, where $\W\in\Real^{d\times d}$, and $\tilde{\X}$ is the augmented design matrix with a column of ones to account for the bias term, $\Xt=[\X,{\bf 1}_n]$. 
The solution to this least squares problem is given by the Wiener-Hopf normal equations, $\W=(\Xt^{\top}\Xt)^{-1}\Xt^{\top}\Y$. The predictions can be expressed as $\hat{\Y}=\Xt\W=\Xt(\Xt^{\top}\Xt)^{-1}\Xt^{\top}\Y={\H}\Y$, where $\mathbf{H}$ which is known as the {\em projection matrix}. Now, by taking the derivative of the predictions $\hat{\Y}$ w.r.t. $\Y$, $\H = \frac{\partial\hat{\Y}}{\partial\Y}$. The $i$-th element of the diagonal of $\H$ is thus given by 
\begin{equation}
\label{eq:cook}    
h_i=\x_i^{\top}(\Xt^{\top}\Xt)^{-1}\x_i,
\end{equation}
and is known as the leverage of the $i$-th observation. 
Similarly, the $i$-th element of the residual vector $\mathbf{e} =\mathbf{y} -\mathbf{\hat {y}} =\left(\mathbf{I} -\mathbf{H} \right)\mathbf{y}$ is denoted by $e_{i}$.

The Cook's distance $D_i$ for observation $\x_i$, $i = 1,\ldots, n$, is defined as the sum of all the changes in the regression model when observation $i$-th is removed from it: 
\begin{equation}
\label{eq:cook0}
    D_i=\dfrac{\sum_{j=1}^{n}\left(\hat{\y}_j-\hat{\y}_{j\backslash i}\right)^2}{d~\text{MSE}^2},
\end{equation}
where $\hat{\y}_{j\backslash i}$ is the fitted response value obtained when excluding $i$, and MSE is the mean-square error of the regression model. Equivalently, it can be expressed using the leverage
\begin{equation}
\label{eq:cook}
    D_i=\dfrac{e_i^2 h_i}{d~\text{MSE}^2(1-h_i^2)}.
\end{equation}
Interestingly, Cook showed that this estimation can be obtained using incremental rank one update of covariances, without even needing to re-compute each models when the $i$-th sample is removed~\cite{Cook77}.

\subsection{Nonlinear randomized Cook's distance}

The application of the Cook's distance can actually consider nonlinear regression to obtain more accurate estimates and hence sharper anomaly detections from the residuals. Note, however, that the predictive function should be fast to train and evaluate so we can plug it in~\eqref{eq:cook0}, or tractable as in the linear case to be able to derive an equivalent equation to~\eqref{eq:cook} but involving a {\em nonlinear leverage function} $h_i'$. 

Kernel methods are well-equipped tools to cope with nonlinear problems~\cite{ShaweTaylor04} while still resorting to linear algebra operations. 
We should note, however, that the extension of the Cook's distance in terms of kernels imposes a large computational cost and resorting to incremental inverses on Gram (kernel) matrices. In this paper, we explore an alternative, simple way: we approximate the nonlinear solution with random Fourier features, which approximate shift-invariant kernels~\cite{Rahimi07nips}, we will refer to our proposed method as the randomized Cook's distance (or RCook) in advance. 

Formally, we now use a linear model expressed on data projected onto $D$ random Fourier features. Let us define a feature map ${\bf z}({\bf x}):\Real^d\to\Comp^D$, {\em explicitly} constructed as ${\bf z}({\bf x}):=[\exp(\imag{\bf w}_1^\top{\bf x}),\ldots,\exp(\imag{\bf w}_D^\top{\bf x})]^\top$, where $\imag=\sqrt{-1}$, 
and ${\bf w}_i \in \Real^{d}$ is randomly sampled from a data-independent distribution~\cite{Rahimi07nips}. The prediction model is now defined as $\hat{\Y}=\Re\{\Z\W\}$, where ${\bf Z}=[{\bf z}_1\cdots{\bf z}_n]^\top\in\Real^{n\times D}$, with the weight matrix $\W\in\Real^{D\times d}$.
The nonlinear randomized leverage of a particular sample is now expressed 
\begin{equation}
h_i'=\z(\x_i)(\Z^{\top}\Z)^{-1}\z(\x_i),
\end{equation}
which is then plugged into~\eqref{eq:cook} owing to the linearity of the model. 
This allows to control the memory and computational complexity explicitly through $D$, as one has to store matrices of $n\times D$ and invert matrices of size $D\times D$ only. It is worth noting that a few number of random Fourier features are needed in practical applications, $D\ll n$. This is not only beneficial in computation time and memory savings, but also has a  regularization effect in the solution.
Table~\ref{space_time_complexity} shows the space and time (computational) efficiency of both methods since $d$ and $D$ are generally less then hundreds.

\begin{table}[h]
\centering
\caption{Space and time complexity for both methods.}
\label{space_time_complexity}
\begin{tabular}{|l|l|l|l|l|l|l|}
\hline\hline
\rowcolor[HTML]{A2B1C5}
\bf{Method} &  $\bf{T}$      & $\bf{C}$   & $\bf{C^{-1}}$    & $\bf{W}$     & $\bf{L}$      & $\bf{ACD}$   \\ \hline\hline
 \rowcolor[HTML]{DBE3EE}\multicolumn{7}{|l|} {\bf{Space}} \\ \hline\hline
Cook   &  -           & -     & $d^2$      & $d^2$    &$n$       & $n$   \\ \hline
R-Cook &  $nD$        & -     & $D^2$      & $D^2$    &$n$       & $n$   \\ \hline\hline
\rowcolor[HTML]{DBE3EE}
\multicolumn{7}{|l|} {\bf{Time}}\\ \hline\hline
Cook     &  -      & $nd^2$   & $d^3$  & $nd^2$  & $nd^2$ & $nd^2$        \\ \hline
R-Cook   &  $ndD$  & $nD^2$  & $D^3$  & $nD^2$  &$nD^2$  & $nD^2$  \\ \hline

\end{tabular}
\vspace*{1pt}
\newline\footnotesize{$T$ is transformation of image into a nonlinear space.}
\newline\footnotesize{$C$ is for covariance matrix and $C^{-1}$ is for its inverse.}
\newline\footnotesize{$W$ is for regression weight and $L$ is for leverage.}
\end{table}


\section{Experiments}
\label{sec:experiments}

\begin{figure}[h!]
\centering
\subfloat[March 2017]{\includegraphics[width=2.55cm]{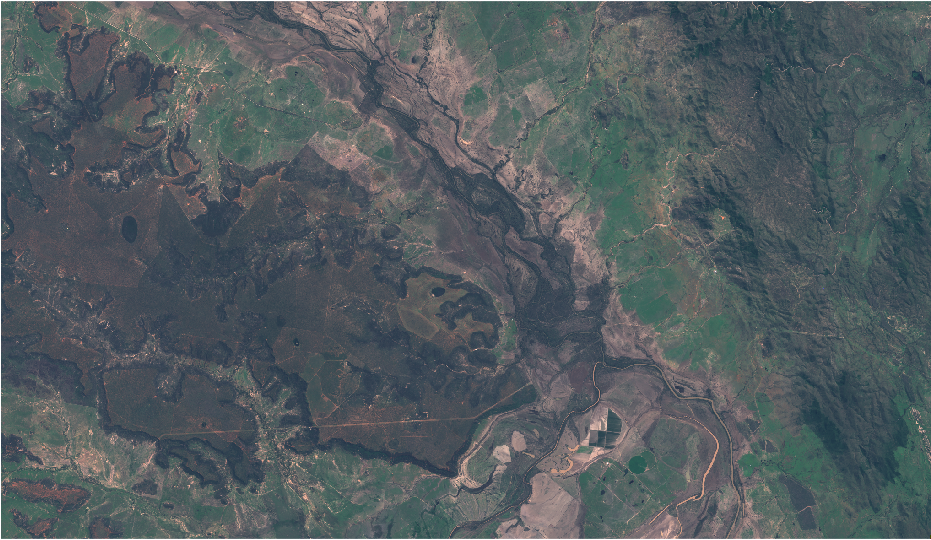}} \hspace{1mm}
\subfloat[May 2017]{\includegraphics[width= 2.55cm]{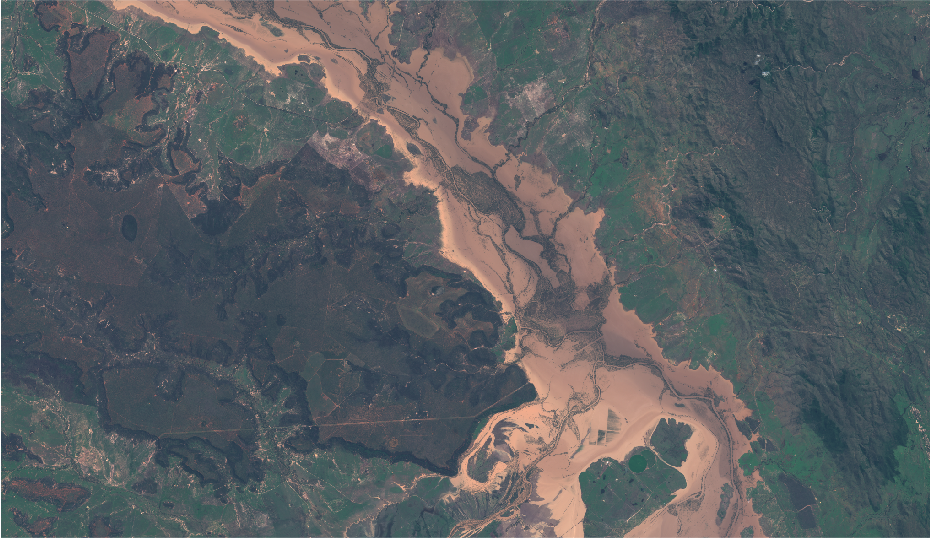}} \hspace{1mm}
\subfloat[Ground Truth]{\includegraphics[width= 2.55cm]{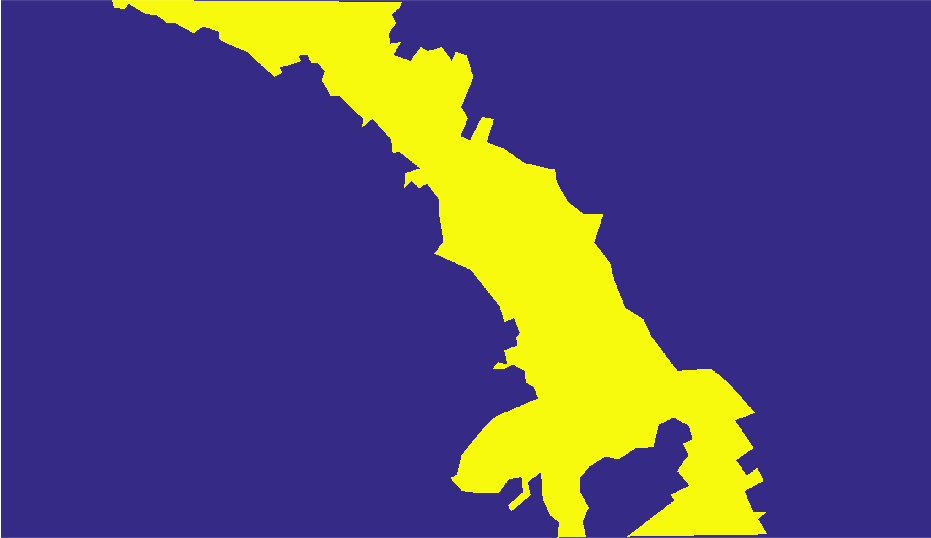}}\\
\vspace{-0.3cm}
\subfloat[Aug 8th 2017]{\includegraphics[width=2.55cm]{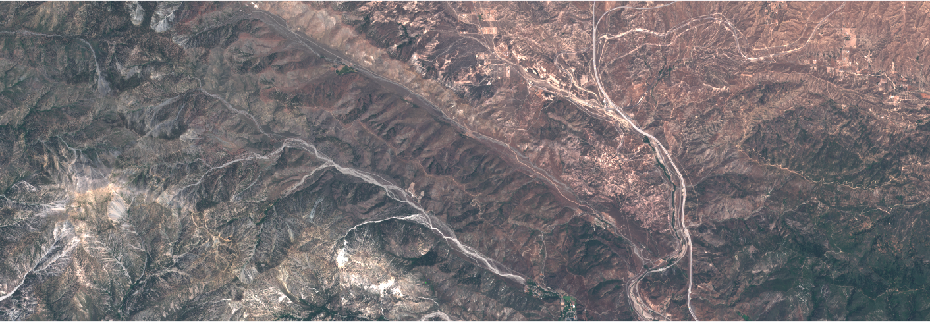}}
\hspace{1mm}
\subfloat[Aug 28th 2017]{\includegraphics[width=2.55cm]{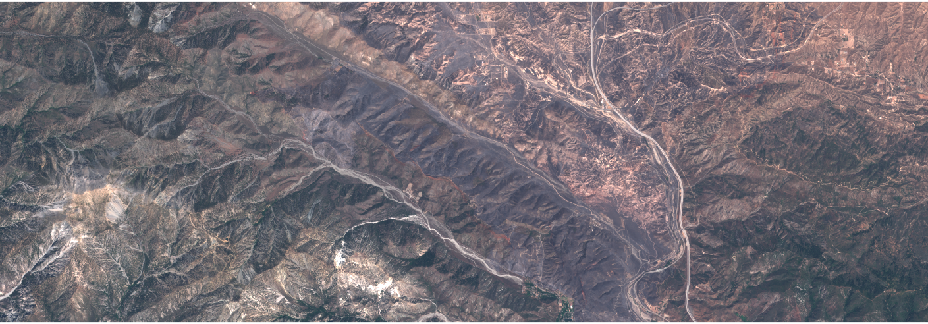}}
\hspace{1mm}
\subfloat[Ground Truth]{\includegraphics[width=2.55cm]{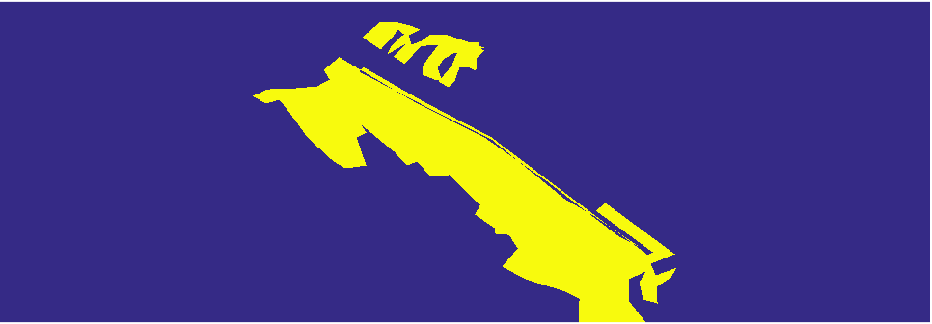}} 
\hspace{1mm}\\
\vspace{-.3cm}
\subfloat[2002]{\includegraphics[width= 2.55cm]{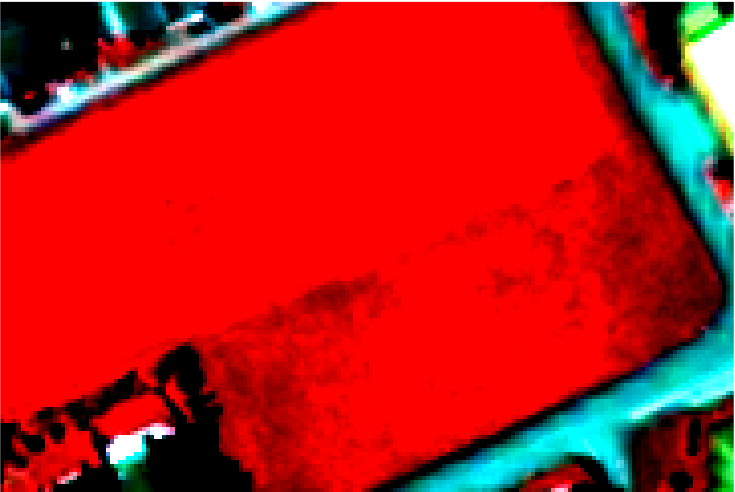}} \hspace{1mm}
\subfloat[2006]{\includegraphics[width= 2.55cm]{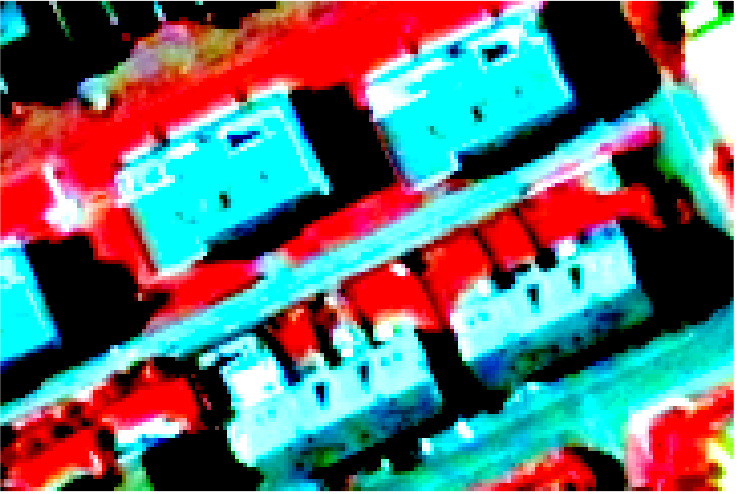}} \hspace{1mm}
\subfloat[Ground Truth]{\includegraphics[width= 2.55cm]{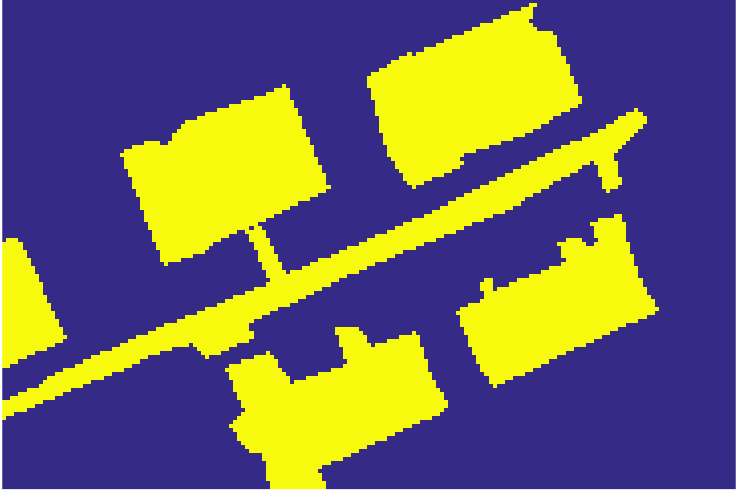}} \hspace{1mm}\\
\vspace{-.3cm}
%
\subfloat[March 2017]{\includegraphics[width=2.55cm]{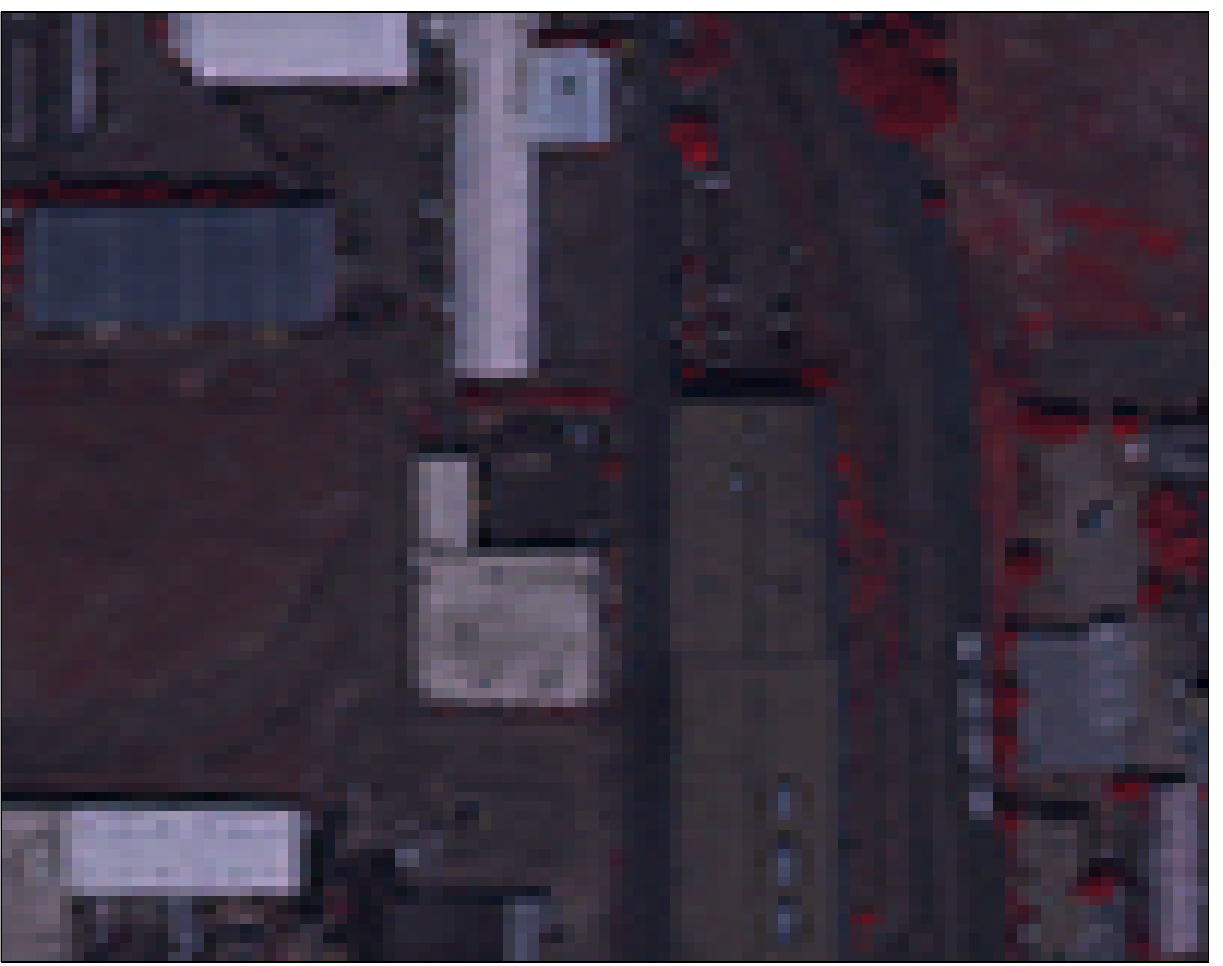}} \hspace{1mm}
\subfloat[May 2017]{\includegraphics[width= 2.55cm]{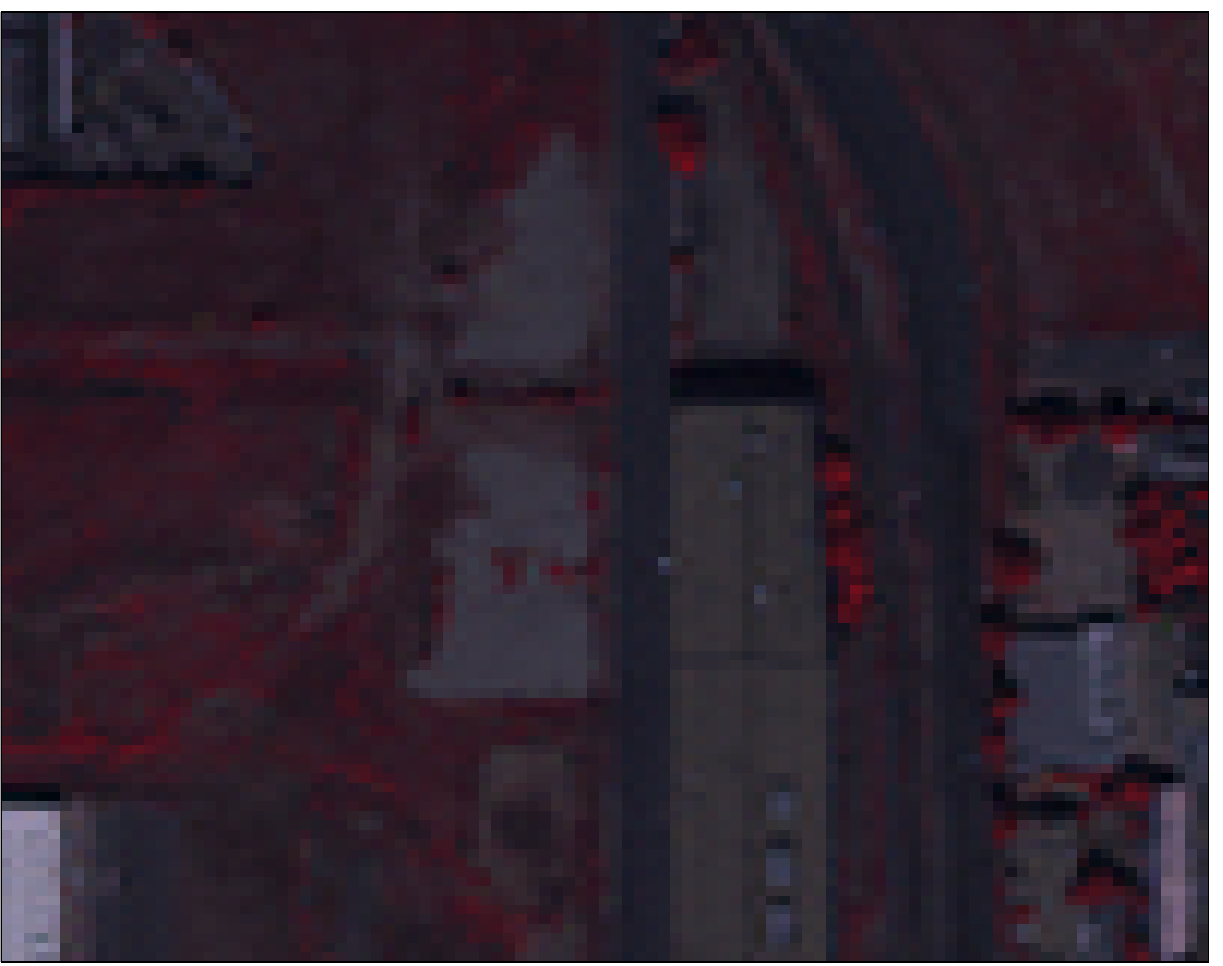}} \hspace{1mm}
\subfloat[Ground Truth]{\includegraphics[width= 2.55cm]{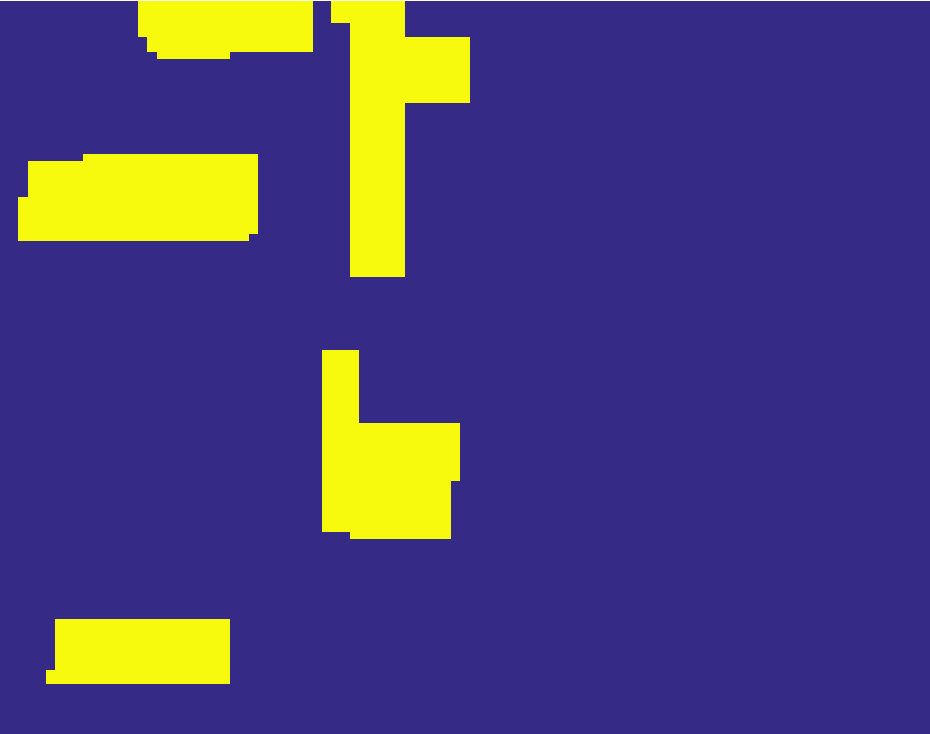}}\\
\vspace{-0.3cm}
\caption{Images with real anomalies. \label{fig:rgb}}
\end{figure}

In order to test the different methods in a common framework, we fix a database formed by four multispectral images. 
We generated the ground truth of each image by photo-interpretation, and consequently we know where the anomalous change is present. All images contain changes of different kind, which allow us to study how the different algorithms perform in a diversity of realistic scenarios.
In Table~\ref{table:database} appears different descriptors of the images in the database, the sensor which has acquired the scene, the number of rows and columns (whose product is the total amount of pixels) and the number of bands of each one. 
Fig.~\ref{fig:rgb} illustrates the RGB composites, the left column are images without anomaly, the middle column is where appears it, and the right column the corresponding ground truth. Fig.~\ref{fig:rgb}(a)-(c) correspond to natural floods caused by Cyclone Debbie in Australia 2017.
In Fig.~\ref{fig:rgb}(d)-(f) appears the consequences of the fire in a mountainous area of California (USA).
Fig.~\ref{fig:rgb}(g)-(i) corresponds to developments in the city of Z\"urich (Switzerland).
And in Fig.~\ref{fig:rgb}(j)-(l) appears an urbanized area in the city of Denver (USA). The Denver image used in the study is a sub-image of the original one.

\begin{table}[h!]
\centering
\caption{Images attributes in the experimentation dataset.}\label{table:database}
\begin{tabular}{ |l|c|c|c|c| } 
\hline\hline
\rowcolor[HTML]{A2B1C5} \bf{Image} & \bf{Sensor} & \bf{Rows} & \bf{Columns}  & \bf{Bands} ($d$)      \\ \hline\hline
Australia  & {Sentinel-2} & 1175 & 2032 & 12 \\\hline
California & {Sentinel-2} & 332 & 964 & 12 \\\hline
{Z\"urich} & Quickbird & 181 & 121  & 4 \\\hline 
Denver & Quickbird & 101 & 101 & 4\\\hline
\end{tabular}
\end{table}


We fix the experimentation setup with $n=10^4$ samples, split into training and test sets, with 50\% each one, and select the best parameters by cross-validation in the training set. 
Nonlinear methods involve the optimization of the $\sigma$ parameter and the $\lambda$ regularization parameter. For both of them, we make a grid search by taking 50 logarithmically-spaced points between [$10^{-5},10^{4}$]. We optimized the different methods by maximizing the area under the curve (AUC) of the receiver operating characteristic (ROC) curve. In the case of RCook, we tested different values for the effective dimensionality parameter, $D$, and finally set $D=100$ which leads to a good performance and a moderate computational execution.

\begin{figure*}
\subfloat[Australia]{\includegraphics[width=.2\textwidth]{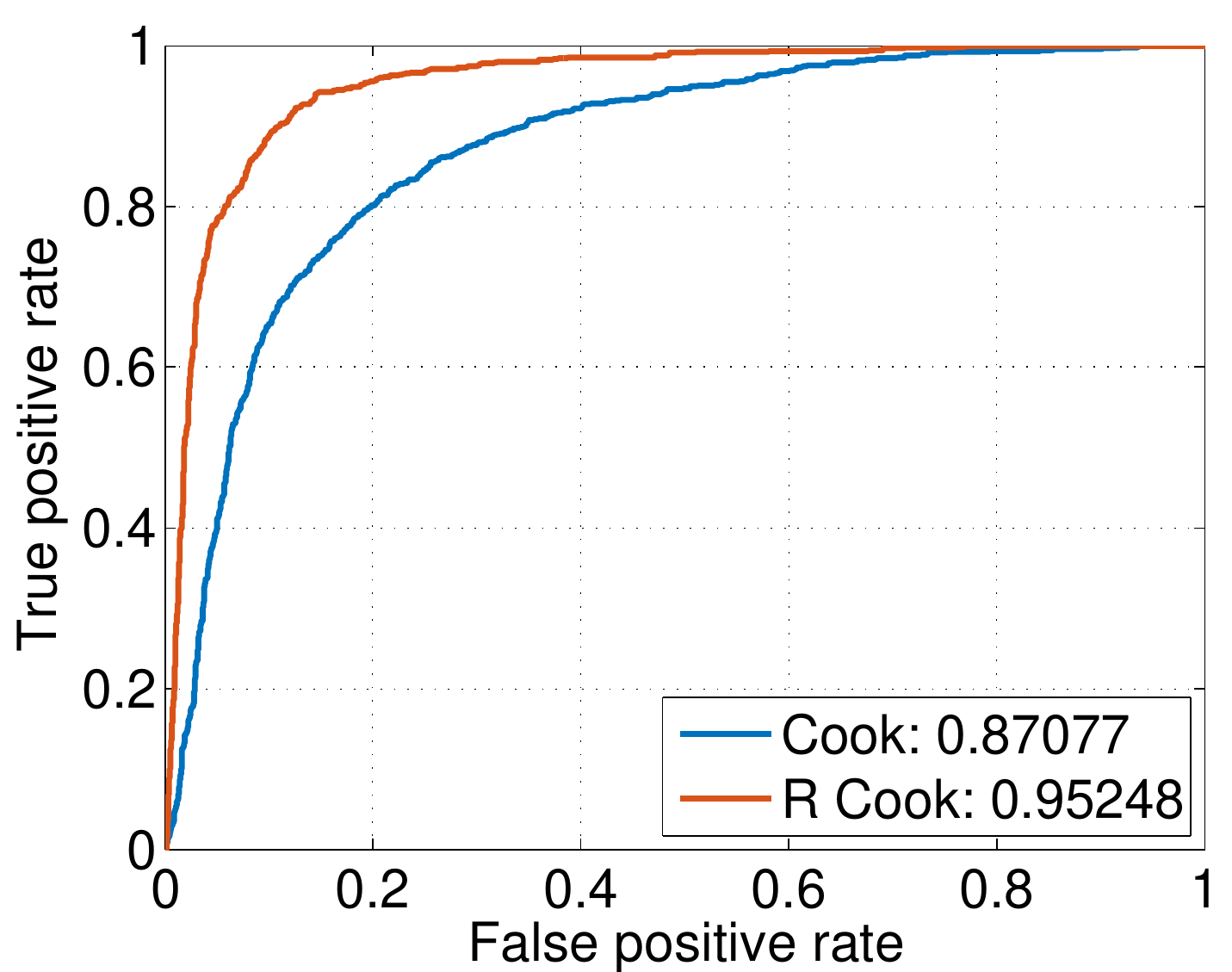}}
\subfloat[California]{\includegraphics[width=.2\textwidth]{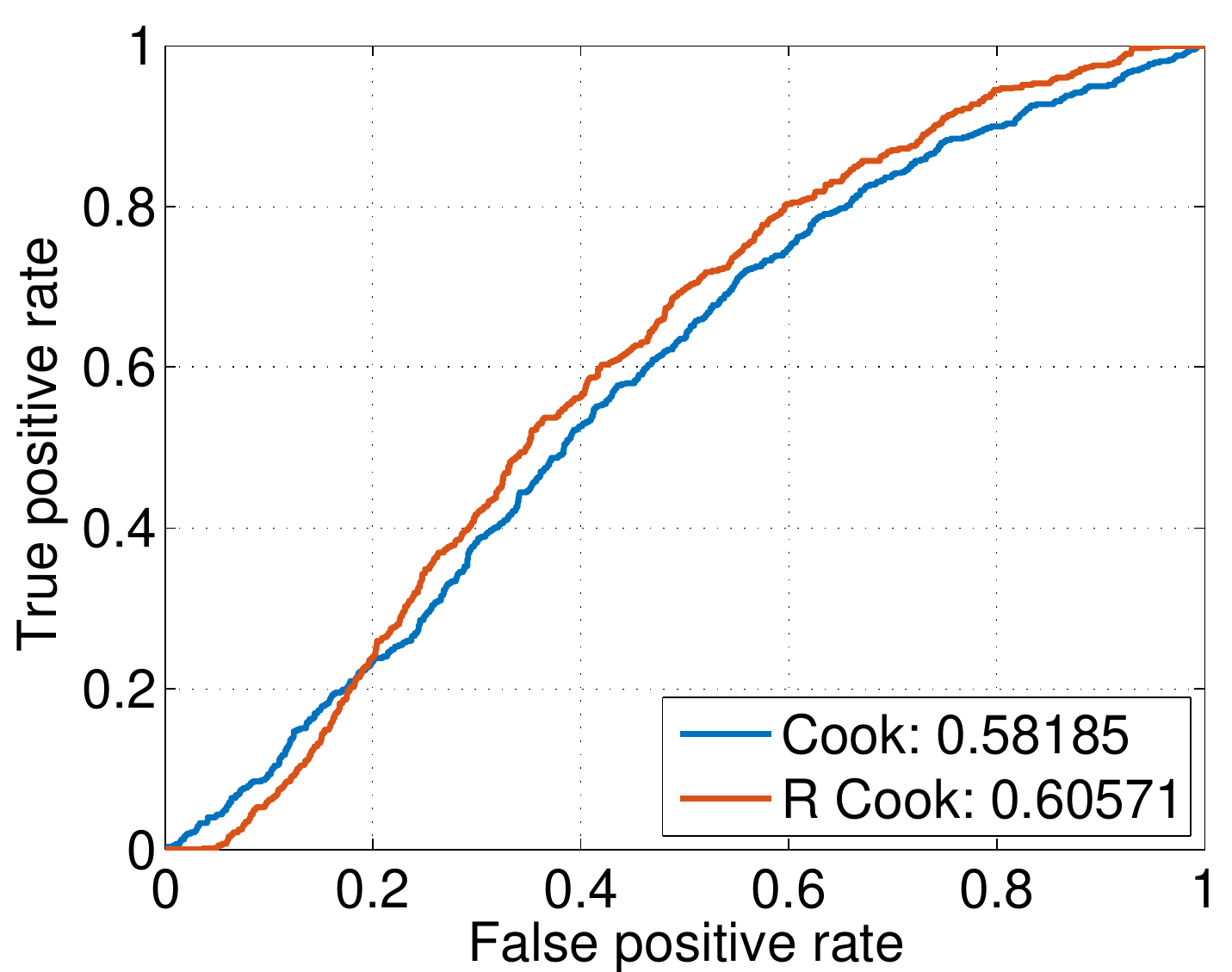}}
\subfloat[Z\"urich]{\includegraphics[width=.2\textwidth]{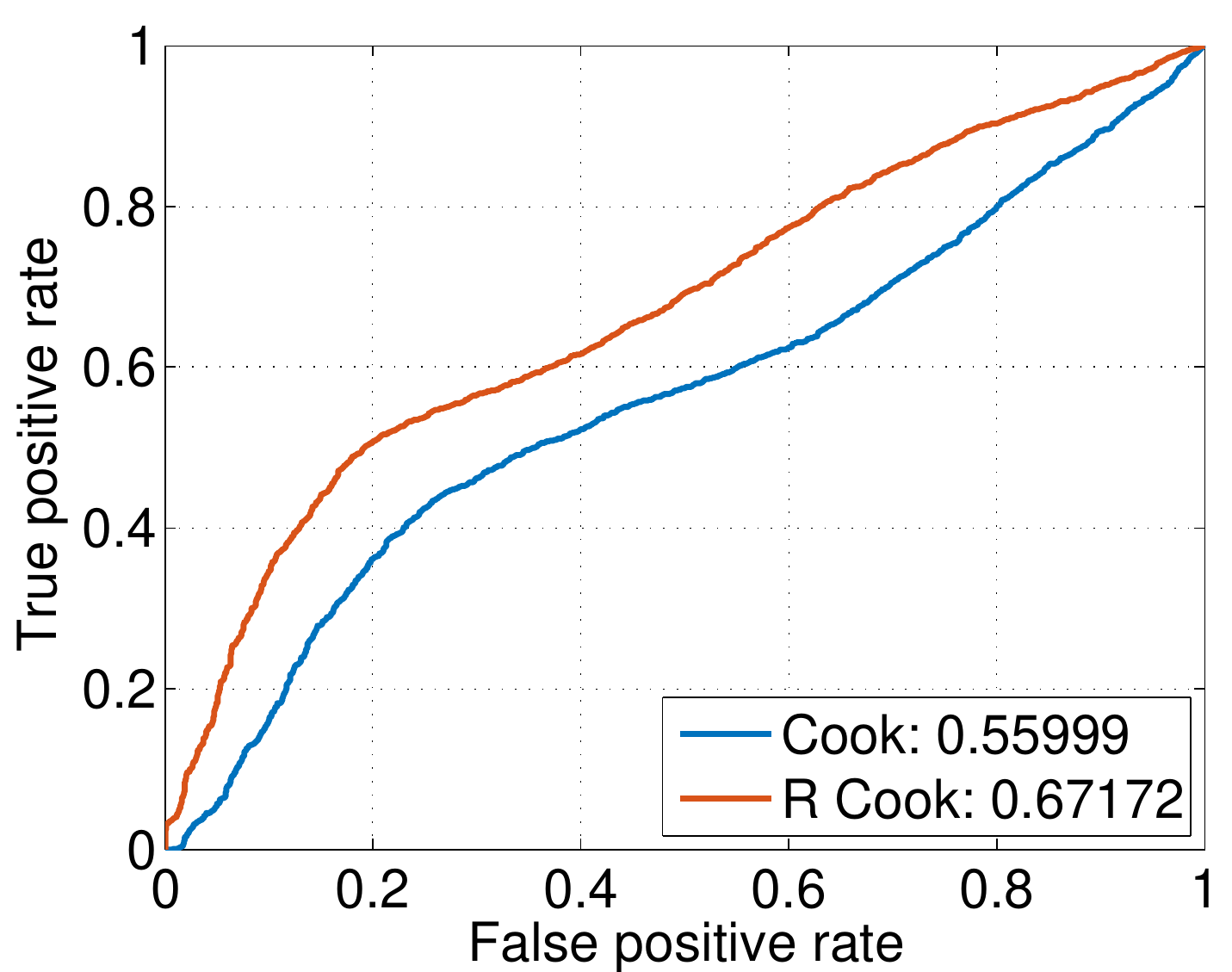}}
\subfloat[Denver ($n_{\text{train}}=5000$)]{\includegraphics[width=.2\textwidth]{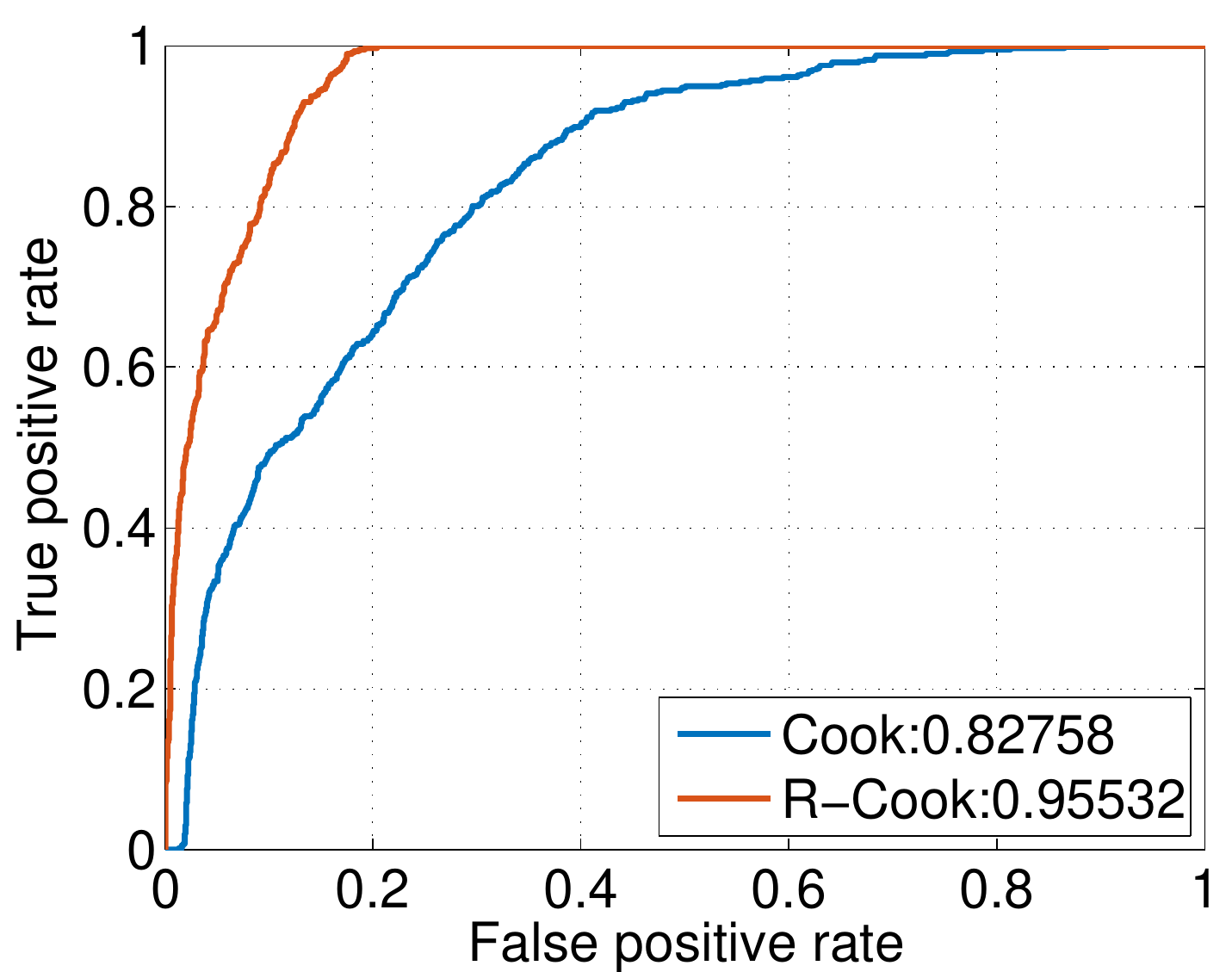}} 
\subfloat[Denver (full)]{\includegraphics[width=.2\textwidth]{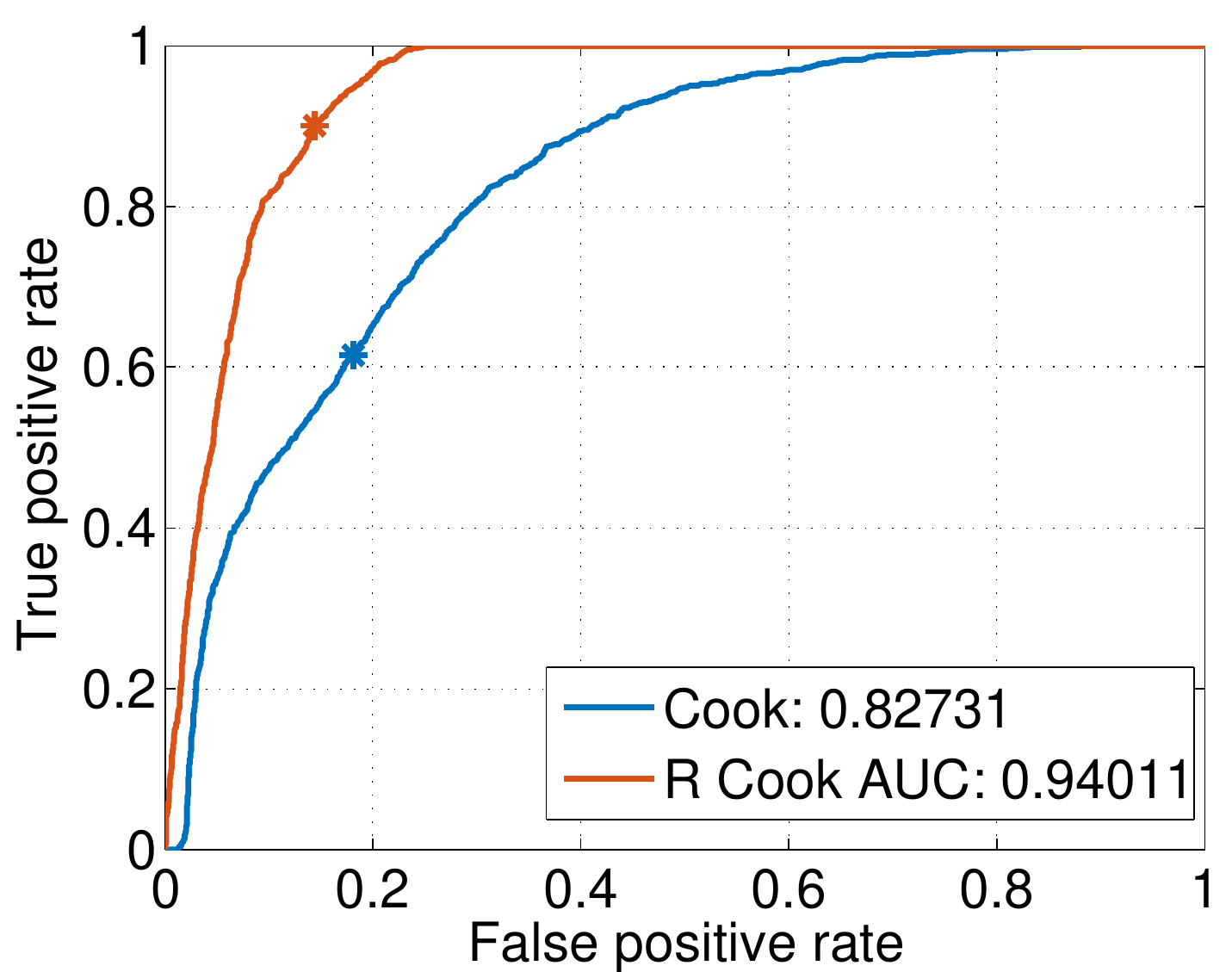}} 
\caption{Linear and randomized Cook's distance ROC curves over different multispectral images. Each legend image contains the AUC values achieved by the methods.}\label{fig:Australia-California}
\end{figure*}

\begin{figure}[!htp]
\includegraphics[width=.32\columnwidth]{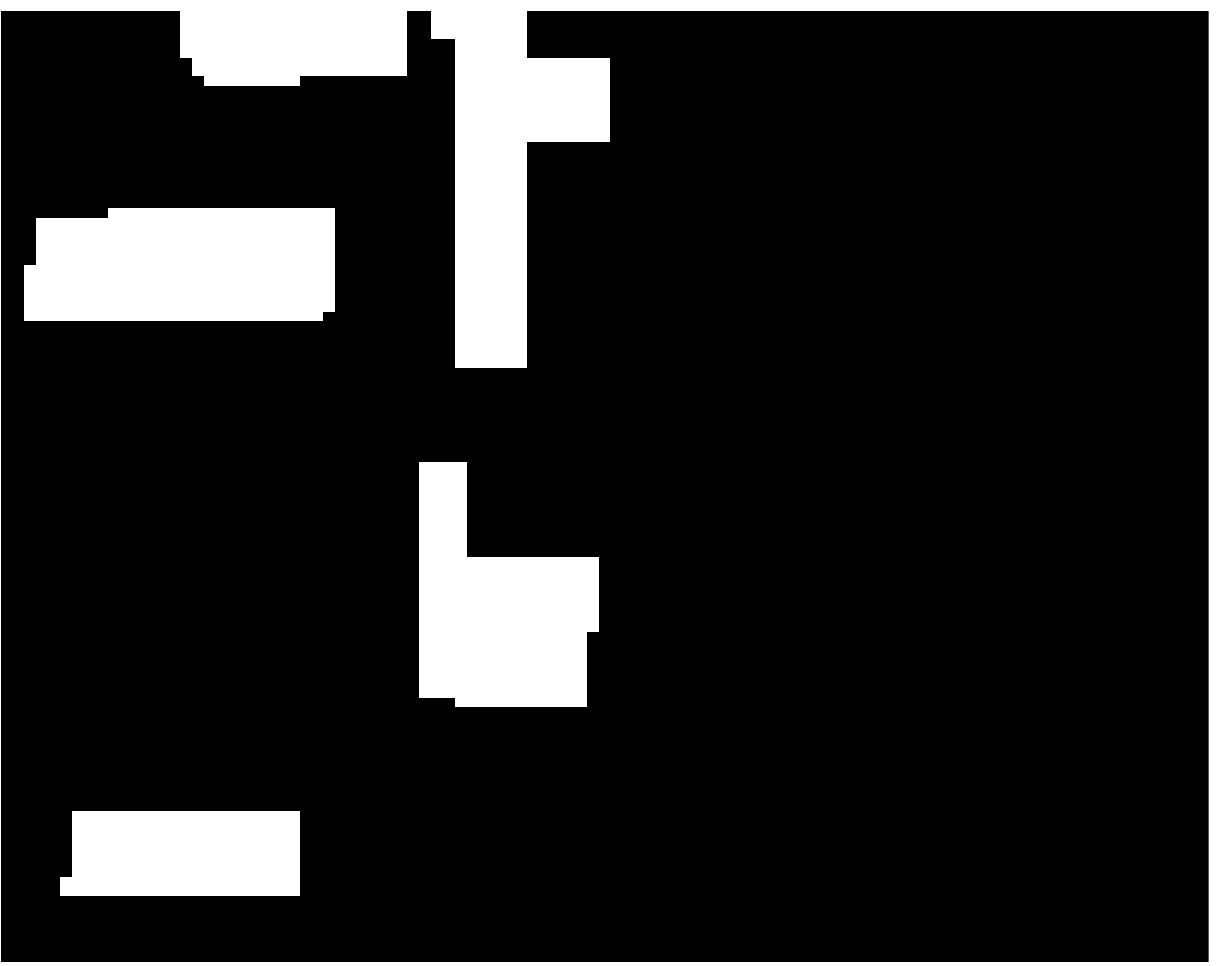}
\includegraphics[width=.32\columnwidth]{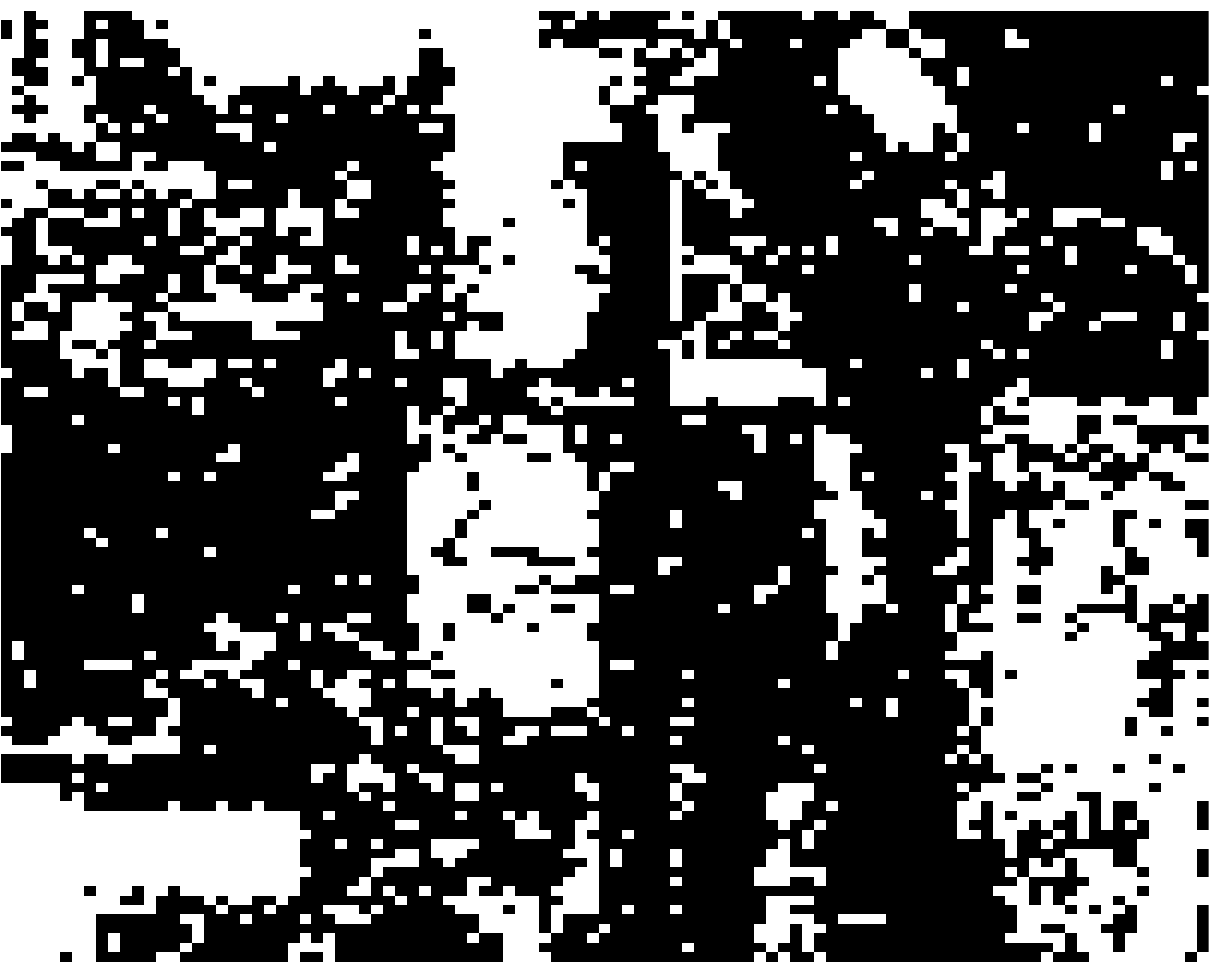}
\includegraphics[width=.32\columnwidth]{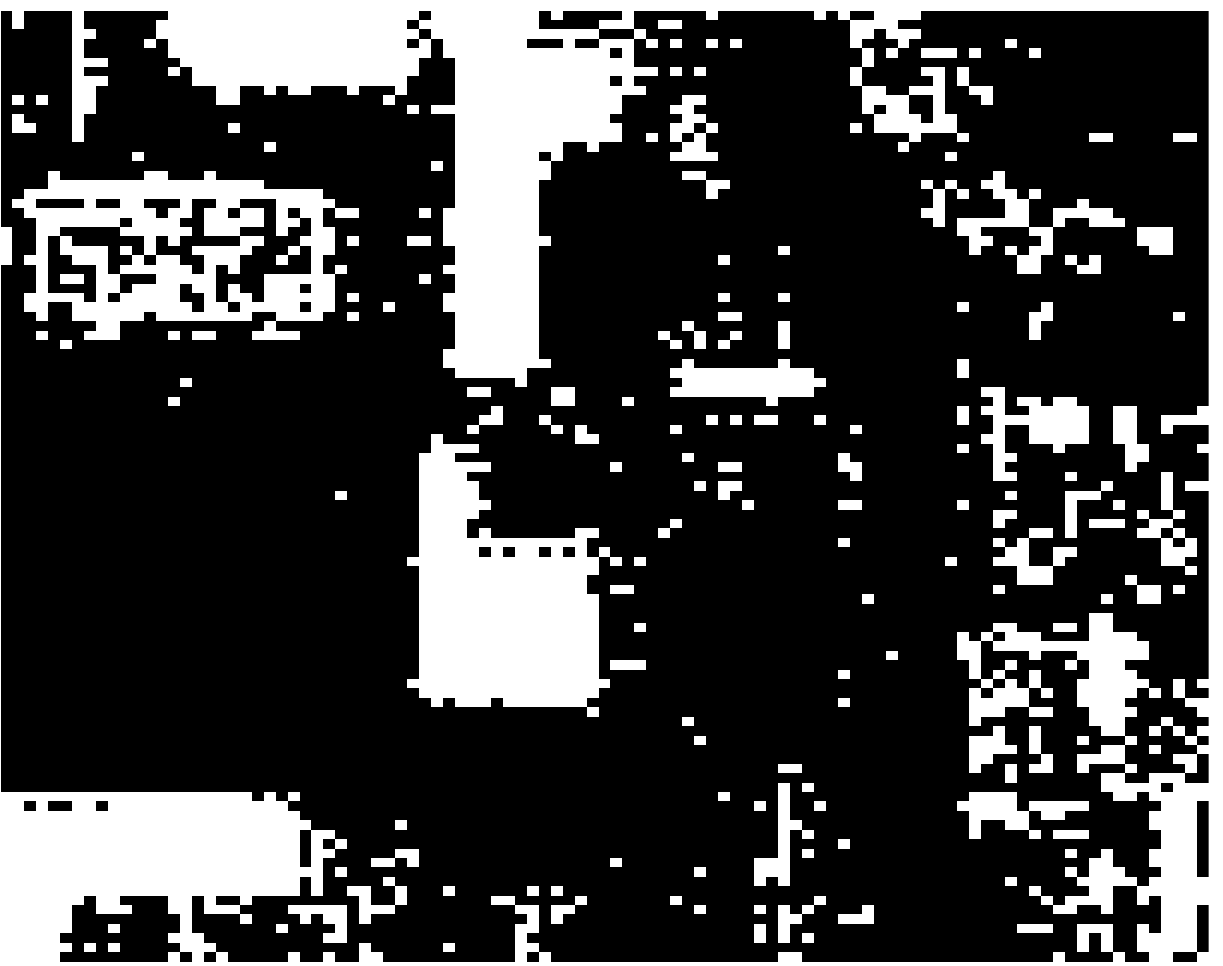}
\caption{Denver anomaly ground truth (left), thresholded Cook distance values (middle) and thresholded RCook distance values.}\label{fig:Denver-classify}
\end{figure}

Figure~\ref{fig:Australia-California}(a)-(d) shows the ROC curves for all the images in the database following the described setup. Each plot legend contains the AUC value of each method. There is a clear advantage of the randomized Cook's distance which outperforms its linear counterpart by uppering the ROC curve and obtaining always a greater AUC value. 
The gain is not noticeable in the case of the California scene, probably due to the complexity involved in the labelling the ground truth and specially in the shadowed parts. More efforts will be taken in this part to attain it. Fig.~\ref{fig:Australia-California}(e) illustrates the results for the Denver image in the full scene. Marked points (nearest to $(0,1)$) in both curves are   selected to set the threshold and classify the full Denver image which appears in Fig.~\ref{fig:Denver-classify}, (left) the ground truth, (middle) result for the linear version which detects more false positives, and (right) the RCook result which can be confirmed visually how allows to obtain a better detection in the whole scene. 

\begin{table}[h!]
\centering
\small
\caption{AUC values for Cook and RCook methods in different images. Both methods are trained with $n=5000$ samples and RCook with $D=100$.}
\label{AUC_table}
\begin{tabular}{ |l|l|l| } 
\hline\hline
\rowcolor[HTML]{A2B1C5} \bf{Scenes}             & \bf{Cook}        & \bf{RCook}         \\ \hline\hline

Australia           &    {\it{0.87}}         &     {\bf 0.95}  
\\\hline
California          & {\it{0.58}}        &       {\bf 0.61}
\\\hline
 Z\"urich             &     {\it{{0.56}}}        &       {\bf{0.67}} 
\\\hline 
Denver & {\it{0.83}} & {\bf 0.95} \\\hline
\end{tabular}
\end{table}

In Table~\ref{AUC_table} we have summarized the AUC values of the linear and RCook over the different images used in the experimentation. As a summary, the randomized Cook distance achieves the best results in all the images compared against its linear counterpart.

\section{Conclusions}
\label{sec:conclusion}

In this work we have presented the randomized Cook's distance. This method uses an explicit nonlinear mapping to transform the data and then follows the standard linear approach. 
Besides, the proposed method is a nonlinear version of the standard Cook distance which allows to find influential points (considered as anomalies) in a scenario following the chronochrome philosophy. 

We have developed an exhaustive statistical experimentation over real data scenes based on the study of the ROC curves and tuned all the methods to achieve maximum performance in the AUC measure. We have shown that the proposed method outperforms the standard linear one in all images.

The proposed methodology has a good generalization behavior, the hyperparameters of the method are optimized through Cross Validation and once them are fixed it can be applied to the whole image. Due to the low computational demand of the RCook method and by following this strategy both open the way to apply over real images of relatively big size. Preliminary results indicates that we can test our RCook method over images around millions of points.

\small
\bibliographystyle{IEEEbib}
\bibliography{bibFile}

\begin{thebibliography}{10}

\bibitem{Lu04}
D.~Lu, P.~Mausel, E.~Brond\'izio, and E.~Moran,
\newblock ``Change detection techniques,''
\newblock {\em Intnl. Jour. Rem. Sens.}, vol. 25, no. 12, pp. 2365--2401, 2004.

\bibitem{Kwon03}
H.~Kwon, S.Z. Der, and N.M. Nasrabadi,
\newblock ``Adaptive anomaly detection using subspace separation for
  hyperspectral images,''
\newblock {\em Opt. Eng.}, vol. 42, no. 11, pp. 3342--3351, 2003.

\bibitem{Mayer03}
R.~Mayer, F.~Bucholtz, and D.~Scribner,
\newblock ``Object detection by using `whitening/dewhitening' to transform
  target signatures in multitemporal hyperspectral and multispectral imagery,''
\newblock {\em IEEE Trans. Geosci. Remote Sens.}, vol. 41, no. 5, pp.
  1136--1142, May 2003.

\bibitem{Arenas13}
J.~Arenas-Garc\'{i}a, K.~Brandt~Petersen, G.~Camps-Valls, and L.~Kai~Hansen,
\newblock ``Kernel multivariate analysis framework for supervised subspace
  learning,''
\newblock {\em IEEE Sig. Proc. Mag.}, vol. 30, no. 4, 2013.

\bibitem{Green88}
A.~A. Green, M.~Berman, P.~Switzer, and M.~D. Craig,
\newblock ``A transformation for ordering multispectral data in terms of image
  quality with implications for noise removal,''
\newblock {\em IEEE Trans. Geosc. Rem. Sens.}, vol. 26, no. 1, pp. 65--74,
  1998.

\bibitem{Nielsen98}
A.~A. Nielsen, K.~Conradsen, and J.~J. Simpson,
\newblock ``Multivariate alteration detection ({MAD}) and {MAF} post-processing
  in multispectral bi-temporal image data: {N}ew approaches to change detection
  studies,''
\newblock {\em Rem. Sens. Env.}, vol. 64, no. 1, pp. 1--19, April 1998.

\bibitem{Schaum97}
A.~Schaum and Stocker A.,
\newblock ``Long-interval chronochrome target detection,''
\newblock in {\em Proc. Int. Symp. Spectral Sens. Res.}, 1997.

\bibitem{Gustau}
Longbotham-Nathan Camps-Valls,
\newblock ``A family of kernel anomaly change detectors,''
\newblock {\em WHISPERS}, pp. 1--4, 2014.

\bibitem{Theiler10}
J.~Theiler, C.~Scovel, B.~Wohlberg, and B.~R. Foy,
\newblock ``Elliptically contoured distributions for anomalous change detection
  in hyperspectral imagery,''
\newblock {\em IEEE Geosc. Rem. Sens. Lett.}, vol. 7, no. 2, pp. 271--275, Apr.
  2010.

\bibitem{REGRESSION1}
F.~J. Anscombe and John~W. Tukey,
\newblock ``The examination and analysis of residuals,''
\newblock {\em Technometrics}, vol. 5, no. 2, pp. 141--160, 1963.

\bibitem{Cook77}
R.~Dennis Cook,
\newblock ``Detection of influential observation in linear regression,''
\newblock {\em Technometrics}, vol. 19, no. 1, pp. 15--18, 1977.

\bibitem{Rahimi07nips}
Ali Rahimi and Benjamin Recht,
\newblock ``Random features for large-scale kernel machines,''
\newblock in {\em Advances in Neural Information Processing Systems 20}, J.~C.
  Platt, D.~Koller, Y.~Singer, and S.~T. Roweis, Eds., pp. 1177--1184. Curran
  Associates, Inc., 2008.

\bibitem{Chronochrome}
Stocker~A. Schaum~A.,
\newblock ``Long-interval chronochrome target detection.,''
\newblock 1997.

\bibitem{ShaweTaylor04}
J.~Shawe-Taylor and N.~Cristianini,
\newblock {\em Kernel Methods for Pattern Analysis},
\newblock Cambridge University Press, New York, NY, USA, 2004.

\end{thebibliography}

\end{document}